\documentclass[10pt, journal, final, twocolumn]{article}

\usepackage{amsmath, amssymb, amsfonts}
\usepackage{cite}
\usepackage{graphicx}
\usepackage{bm} 
\usepackage{stfloats}
\usepackage{hyperref}

\begin{document}

\title{A Unified Formula for Affine Transformations between Calibrated Cameras}

\author{Levente~Hajder
\thanks{L. Hajder is with the Geometric Computer Vision Group, E\"otv\"os Lor\'and University (ELTE), Budapest, Hungary. Email: \textit{hajder@inf.elte.hu}, \url{https://cv.inf.elte.hu}}}

\maketitle

\begin{abstract}
In this technical note, we derive a closed-form expression for the affine transformation mapping local image patches between two calibrated views. We show that the transformation is a function of the relative camera pose, the image coordinates, and the local surface normal.
\end{abstract}

\section{Introduction}
The relationship between local image deformations and scene geometry is a cornerstone of computer vision. The aim of this document is to show how a local affine transformations can be given for a calibrated stereo setup if the implicit parameters of a tangent surface are known.

\section{The Unified Affine Formula}
Let the camera motion be represented by $(\mathbf R, \mathbf{t})$, aka. relative pose between the views, where  $\mathbf R \in SE(3)$ is the rotation, and the 3D vector $\mathbf t=\left[ t_x \quad  t_y \quad t_z  \right]^T$ is the translation between the images.

In the 3D world, we observe a spatial point $\mathbf X$. It is on a tangent surface, for which the normal is denoted by $\mathbf n$. The two cameras take pictures, the projections of $\mathbf X$ are $\mathbf p_1$ and $\mathbf p_2$ in the images. We assume that the cameras are pin-hole, the pixel coordinates are normalized by the inverse of the intrinsic matrices $\mathbf K_1$ and $\mathbf K_1$. The local affine transformation between the projected patches are denoted by matrix $\mathbf A$, its size is $2 \times 2$.

\begin{figure}[htbp]
    \centering
    \includegraphics[width=0.45\textwidth]{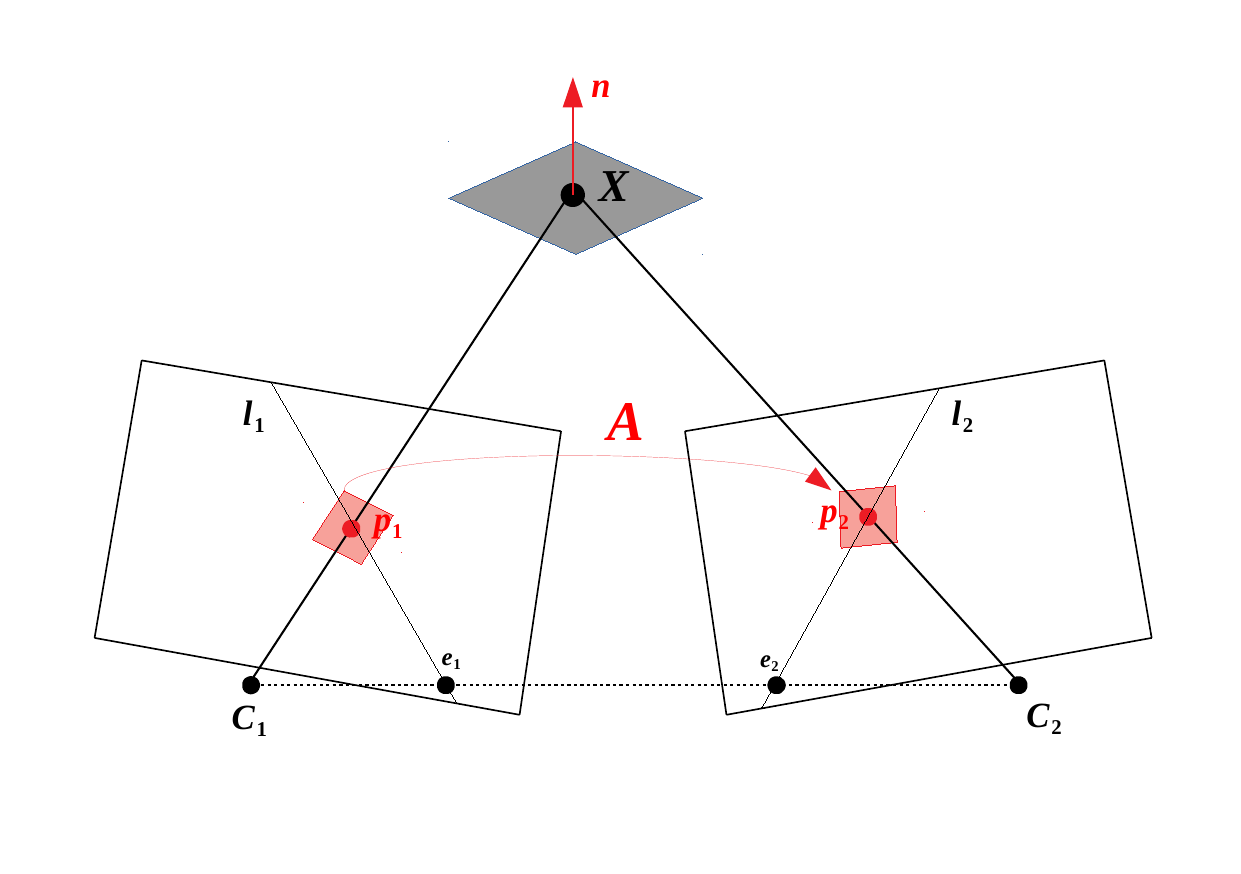}
    \caption{The 3D geometry of the affine transformations. Projected patches and surface normals are highlighted as they are in a strong relationship.}
    \label{fig:problem}
\end{figure}

The full problem is visualized in Fig.~\ref{fig:problem}. The epipoles $\mathbf e_1$ and $\mathbf e_2$ and corresponding epipolar lines $\mathbf l_1$ and $\mathbf l_2$ are also drawn for the sake of completeness. The plot highlights that the local affine transformation $\mathbf A$ is strongly influenced by the surface normal $\mathbf n$.

As it is well-known in stereo vision~\cite{Hartley2003}, a homography $\mathbf H$ maps the projected point of a planar surfaces from one pin-hole camera image to another. The homography $\mathbf H$ can be written~\cite{Faugeras1988,Malis2007} as

\begin{equation*}
\mathbf H=\mathbf R-\frac{\mathbf t \mathbf n^{T}}{d}.
\end{equation*}

Substituting the elements of the matrix and vectors gives the following relationship:

\begin{equation}
\mathbf H=\left[\begin{array}{ccc}
R_{11}-\frac{t_{x}n_{x}}{d} & R_{12}-\frac{t_{x}n_{y}}{d} & R_{13}-\frac{t_{x}n_{z}}{d}\\
R_{21}-\frac{t_{y}n_{x}}{d} & R_{22}-\frac{t_{y}n_{y}}{d} & R_{23}-\frac{t_{y}n_{z}}{d}\\
R_{31}-\frac{t_{z}n_{x}}{d} & R_{32}-\frac{t_{z}n_{y}}{d} & R_{33}-\frac{t_{z}n_{z}}{d}
\end{array}\right].
\end{equation}
if $R_{ij}$ denotes the element of $\mathbf R$ in the i-th row and j-th columns. The planar surface is represented by normal $\mathbf n =\left[ n_x \quad  n_y \quad n_z  \right]^T$ and $d$, the latter one is the distance of the plane with respect to the first image,

As it is written in our previous paper~\cite{Barath2017}, the affine parameters can be obtained by taking the partial derivatives of the coordinates in the second image w.r.t those in the first one.
For example,

\begin{equation}
a_{11}=\frac{H_{11}-H_{31}u_2}{\left(H_{31}u_1+H_{32}v_1+H_{33}\right)}=\frac{b_{11}}{s},
\end{equation}

if the following notations are introduced:

\begin{eqnarray}
b_{11}=&R_{11}-\frac{t_{x}n_{x}}{d}-u_2 \left(R_{31}-\frac{t_{z}n_{x}}{d}\right),
\\
=&R_{11}-u_2 R_{31}-\frac{n_{x}}{d}\left(t_{x}-u_2 t_{z}\right)
\end{eqnarray},

and

\begin{eqnarray*}
s=&H_{31}u_1+H_{32}v_1+H_{33}
\\
=&\left(R_{31}-\frac{t_{z}n_{x}}{d}\right)u_1+\left(R_{32}-\frac{t_{z}n_{z}}{d}\right)v_1+\left(R_{33}-\frac{t_{z}n_{y}}{d}\right)
\\
=&\mathbf r_{3}^{T} \mathbf p_1+\frac{t_{z}}{d}\mathbf n^{T}\mathbf p_1=\left(\mathbf r_{3}^{T}+\frac{t_{z}}{d} \mathbf n^{T}\right) \mathbf p_1,
\end{eqnarray*}

where the vector $\mathbf p_i$, $i \in \{1,2\}$, contains the coordinates of the corresponding point in the first image in homogeneous form, i.e., $\mathbf p_i=[u_i\quad v_i\quad1]^{T}$, and the vector $\mathbf r_{3}$ denotes the third row of the rotation matrix $\mathbf R$.

Similarly,

\begin{equation*}
a_{12}=\frac{b_{12}}{s},\quad a_{21}=
\frac{b_{21}}{s},\quad a_{22}=\frac{b_{22}}{s},
\end{equation*}

if

\begin{eqnarray*}
b_{12}=&h_{12}-h_{32}u_2
\\
=&R_{12}-\frac{t_{x}n_{y}}{d}-u_2\left(R_{32}-\frac{t_{z}n_{y}}{d}\right)
\\
=&R_{12}-u_2 R_{32}-\frac{n_{y}}{d}\left(t_{x}-u_2 t_{z}\right),
\end{eqnarray*}
and
\begin{eqnarray*}
b_{21}=&h_{21}-h_{31}v_2 
\\
=&R_{21}-\frac{t_{y}n_{x}}{d}-v_2\left(R_{31}-\frac{t_{z}n_{x}}{d}\right)
\\
=&R_{21}-v_2 R_{31}-\frac{n_{x}}{d}\left(t_{y}-v_2 t_{z}\right).
\end{eqnarray*}

Finally,

\begin{eqnarray*}
b_{22}=&h_{22}-h_{32}v_2 
\\
=&R_{22}-\frac{t_{y}n_{y}}{d}-v_2 \left(R_{32}-\frac{t_{z}n_{y}}{d}\right)
\\
=&R_{22}-v_2 R_{32}-\frac{n_{y}}{d}\left(t_{y}-v_2 t_{z}\right).
\end{eqnarray*}

By substituting these back into the complete affine transformation we get Equation~\ref{eq:aff_main}.

\begin{figure*}[b]
\hrulefill
\begin{eqnarray}
\mathbf A=&\frac{1}{s}\left[\begin{array}{cc}
b_{11} & b_{12}\\
b_{21} & b_{22}
\end{array}\right] \nonumber
\\
=&\frac{1}{s}\left[\begin{array}{cc}
R_{11}-u_2 R_{31}-\frac{n_{x}}{d}\left(t_{x}-u_2 t_{z}\right) & R_{12}-u_2 R_{32}-\frac{n_{y}}{d}\left(t_{x}-u_2 t_{z}\right)\\
R_{21}-v_2 R_{31}-\frac{n_{x}}{d}\left(t_{y}-v_2 t_{z}\right) & R_{22}-v_2 R_{32}-\frac{n_{y}}{d}\left(t_{y}-v't_{z}\right)
\end{array}\right] \nonumber
\\
=&\frac{1}{s}\left(\left[\begin{array}{cc}
R_{11} & R_{12}\\
R_{21} & R_{22}
\end{array}\right]-\left[\begin{array}{c}
u_2\\
v_2
\end{array}\right]\left[\begin{array}{cc}
R_{31} & R_{32}\end{array}\right]-\frac{1}{d}\left[\begin{array}{c}
t_{x}-u_2 t_{z}\\
t_{y}-v_2 t_{z}
\end{array}\right]\left[\begin{array}{cc}
n_{x} & n_{y}\end{array}\right]\right) \label{eq:aff_main}
\end{eqnarray}
\hrulefill0
\end{figure*}

In the obtained form, the affine transformation can be divided into the sum of three $2 \times 2$ matrices, the first of which is the upper left block of the rotation matrix, while the second and third elements are each a dyad. Interestingly, the third contains the first two coordinates of the normal vector, but this can be misleading because the divisor $s$ is a function of all three elements of the normal vector.


\section{Validation: a Special Case, the Standard Stereo}
Assuming $R = \mathbf{I}$ and $\mathbf{t} = [t_x, 0, 0]^{T}$ gives the standard stereo setup for which $t_x$ represents the baseline. Then

\begin{eqnarray*}
\mathbf A=\frac{1}{s}\left(
\left[\begin{array}{cc}
1 & 0\\
0 & 1
\end{array}\right]- \frac{1}{d}
\left[\begin{array}{c}
t_{x}\\
0
\end{array}\right]\left[\begin{array}{cc}
n_{x} & n_{y}\end{array}\right]\right),
\\
s=\mathbf p^{T}\left(\mathbf r_{3}-\frac{t_{3}}{d}\mathbf n\right)=\mathbf p^{T} \mathbf r_{3}=1
\end{eqnarray*}

Thus,

\begin{eqnarray*}
\mathbf A=\left( \left[\begin{array}{cc}
1 & 0\\
0 & 1
\end{array}\right]- \frac{1}{d} \left[\begin{array}{c}
t_{1}\\
0
\end{array}\right]\left[\begin{array}{cc}
n_{x} & n_{y}\end{array}\right] \right)=
\\
\left[\begin{array}{cc}
1-\frac{t_{1}n_{1}}{d} & --\frac{t_{1}n_{1}}{d}\\
0 & 1
\end{array}\right]
\end{eqnarray*}

The plane is given in implicit form, therefore
\begin{equation*}
n_x x+ n_y y +n_z z +d =0 \rightarrow d= -\left( n_x x+ n_y y +n_z z \right)
\end{equation*}

Therefore,
\begin{eqnarray*}
\mathbf A=
\left[\begin{array}{cc}
1+\frac{t_{1}n_{1}}{n_x x+ n_y y +n_z z} & \frac{t_{1}n_{1}}{n_x x+ n_y y +n_z z}\\
0 & 1
\end{array}\right]
\end{eqnarray*}

This is the same results as published in~\cite{Kariko2025}, the only difference is that the cameras are calibrated here.


\section{Conclusion and Future Work}

In this technical note, we have presented a unified closed-form expression for the local affine transformation between two calibrated views. The proposed Equation ~\ref{eq:aff_main} encapsulates the complex relationship between relative camera pose, image coordinates, and local surface geometry into a single, structured framework. By decomposing the transformation into a sum of three 2×2 matrices, we have provided a clear geometric interpretation of how each component—rotation, translation, and surface orientation—contributes to the final image deformation.

The primary strength of this unified formula lies in its generality, which allows for the effortless derivation of various special cases that were previously treated as separate problems in the literature. While this work validated the formula through the standard stereo setup, it serves as a "parent equation" for numerous other scenarios. Future work will focus on systematically deriving and analyzing these special cases, including:

\begin{itemize}
    \item Planar motion models, where the camera movement is restricted to a specific 2D plane.

    \item Pure translation and pure rotation scenarios, which simplify the dyadic components of the formula.

    \item Small-baseline approximations, providing links to traditional optical flow and scene flow formulations.
\end{itemize}

By providing a consistent mathematical foundation, this work aims to simplify the development of future geometric algorithms in computer vision, ranging from robust feature tracking to direct 3D reconstruction.

\bibliographystyle{ieeetr}
\bibliography{refs}

\end{document}